# SURFACING SEMANTIC ORTHOGONALITY ACROSS MODEL SAFETY BENCHMARKS: A MULTI-DIMENSIONAL ANALYSIS

Jonathan Bennion [1], Shaona Ghosh [2], Mantek Singh [3], Nouha Dziri [4]

[1] The Objective AI, USA
[2] Nvidia, USA
[3] Google, USA
[4] Allen Institute for AI (AI2), USA

## ABSTRACT

*Various AI safety datasets have been developed to measure LLMs against evolving interpretations of harm. Our evaluation of five recently published open-source safety benchmarks reveals distinct semantic clusters using UMAP dimensionality reduction and k-means clustering (silhouette score: 0.470). We identify six primary harm categories with varying benchmark representation. GretelAI, for example, focuses heavily on privacy concerns, while WildGuardMix emphasizes self-harm scenarios. Significant differences in prompt length distribution suggests confounds to data collection and interpretations of harm as well as offer possible context. Our analysis quantifies benchmark orthogonality among AI benchmarks, allowing for transparency in coverage gaps despite topical similarities. Our quantitative framework for analyzing semantic orthogonality across safety benchmarks enables more targeted development of datasets that comprehensively address the evolving landscape of harms in AI use, however that is defined in the future.*

## 1. INTRODUCTION

Differentiated safety interpretations in AI history have led to varied safety evaluation frameworks that have aimed to reduce instances of undesired behavior in language models globally [1], and since these definitions are different to each culture and time, they become collectively opaque [2]. Early AI safety datasets focused on bias evaluation [3], but recent attention has shifted to adversarial robustness and preventing LLM jailbreaks [4]. While general LLM evaluations emphasize depth of intelligence against pre-determined problems to solve [2], the depth has allowed for ample bias to effect the results [5]. Our breadth-based analysis of safety-focused benchmarks reveals coverage gaps and redundancies, highlighting a wider range of failure modes that is still underexplored [6] and improves objectivity in comparing benchmark themes [7]. Using a clustering methodology that maximizes distances between similar semantic meaning based on their encoded numerical values from an embedding model [8], this systematic analysis promotes transparency while reducing the misrepresentation of AI capability improvements known as "safetywashing" [2]. Since these datasets are critical for use as ground truth to measure how any language model may allow for harmful acts, such as content that exacerbates self harmonline [9], the similarities and differences allow for gaps to be addressed in research for future benchmarks.

## 2. RELATED WORKS

The safety benchmarks used in this comparison represent the most recent open-source contributions to AI safety research. AEGIS 2.0 targets commercial use cases, focusing on critical safety concerns in human-LLM interactions and allowing annotators to provide free text input for unclassified risks [9]. WildGuardMix integrates both synthetic and human data with diverse perspectives to ensure objectivity when evaluating novel safety risks, adversarial jailbreaks, and cultural contexts [10]. BeaverTails aligns human preferences within a 14-category safety taxonomy to enhance ethical alignment of LLMs [11]. GretelAI uniquely focuses on evaluating synthetic data generation with privacy guarantees, emphasizing statistical fidelity while preserving differential privacy [12]. MLCommons' AILuminate provides standardized evaluations across twelve hazard categories with over 24,000 test prompts aimed at industrial use [13]. The differences and similarities among these datasets highlight potential gaps in resources available to the open-source community for measuring and adversarial fine-tuning of models to mitigate undesirable language model behavior.

## 3. METHODOLOGY

We compared the semantic differences and similarities of datasets after concatenating each and analyzing the numerical values generated by an embedding model for each prompt string. We used an unsupervised machine learning approach to form clusters of similar semantic values, after cleaning the data and determining the minimum sample size needed for insights. Sample size aimed to measure for categorical frequency differences within harm clusters across datasets was calculated on one iteration for a maximum cluster number of 15 to adhere to the most recent taxonomy [9], and used established research that k means accuracy depends on adequate representation of natural category distributions [14], while targeting an effect size of 0.5, which is presumed to be large enough to have effect on future research [2].

Variables optimized for silhouette score in the full dataset included at least one relevant variant of an embedding model, dimensionality reduction technique, key hyperparameters, distance metrics, and number of clusters, as shown in Figure 1.

After optimizing parameters for defined clusters, we used prompt values at each centroid to infer labels for relevant harm from another language model, then visualize the semantic space differences between benchmarks.

Further analysis of prompt length distributions by benchmark (and by harm by benchmark) illustrate how any potential differences in data collection could contribute the results, since this is the most important aspect to consider for adding context to the results [15].

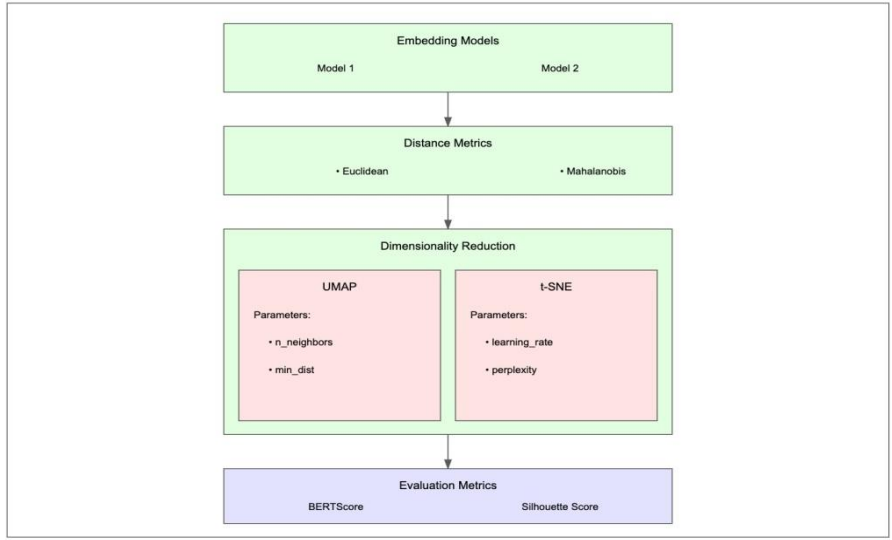

Fig.1. Experimental methodology for clustering optimization configurations.This compared Euclidean and Mahalanobis distance metrics, across two embedding models, followed by dimensionality reduction. Dimensionality reduction included Uniform Manifold Approximation and Projection (UMAP), optimizing *n neighbors* as the most consequential hyperparameter, and t-distributed Stochastic Neighbor Embedding (tSNE), optimizing *perplexity* as the most consequential hyperparameter. Cluster quality was evaluated using Silhouette score as well as processingspeed for scaling (BERTScore was attempted but was not designed to measure clusters and did not differentiate results between configurations).

### 3.1. Sample Size Calculation and Outlier Removal

We aimed to create clusters that maximize silhouette scores for categorical frequency counts between benchmarks, based on prior research indicating these scores reflect internal frequency count differences [14]. Additionally, research indicates that frequency differences in AI safety datasets are insignificant below an effect size of 0.5 [2], supporting its use in Cohen's sample size formula for detecting statistical differences [16], defined below in Equation 1.

$$n_{kmeans} = \frac{2(z_\alpha + z_\beta)^2}{d^2} \tag{1}$$

We also calculated the minimal sample size for each benchmark using past research methods [17], assuming 15 clusters to align with the 12 core harm categories from Aegis 2.0 [9], plus 20 percent for broader semantic coverage [18]. This is illustrated in Equation 2, with $k_{max}$= 15 clusters and $b$ = 5 benchmarks.

$$n_{total}= (max(n_{kmeans}) \times (k_{max} \times 1.2) \times b) \tag{2}$$

Assumptions for this sample size are in Table 1. To enable comparative analysis across all five benchmarks and allow for smaller benchmark sizes, we selected an 85% confidence level ($\alpha$ = 0.15). This trade-off increases Type I error risk for broader dataset inclusion, prioritizing exploratory breadth over strict significance thresholds [16]. Research indicates that model misuse or safety harms in AI use occur frequently enough [2] for this significance level to be acceptable. Based on this, we calculated a cluster sample size of 109, resulting in a total required sample size of 8,175 across all five benchmarks to enable meaningful cluster differentiation.

Table 1. Sample Size Assumptions for Detecting Differences in Harm Categories Per Benchmark

| Parameter | Value |
|---|---|
| Effect size ($d$) | 0.5 |
| Statistical power ($\beta$) | 0.8 |
| Significance level ($\alpha$) | 0.15 |
| Sample size per cluster ($n_{kmeans}$) | 109 |
| Maximum clusters ($k_{max}$) | 15 |
| Number of benchmarks ($b$) | 5 |
| Required sample size per benchmark | 1,635 |
| Total sample size across all benchmarks | 8,175 |

We used IQR bounds to compare prompt length distributions for outlier removal, flagging values outside the typical spread of 50% of the data [20], as shown in Equation 3.

$$[Q_1 - 1.5 \times IQR, Q_3 + 1.5 \times IQR] \tag{3}$$

This was compared against the same distribution using z-score thresholds for outliers, for comparison [19], as shown in Equation 4, where values beyond 3 standard deviations of the mean are considered outliers.

$$\left| \frac{x - \mu}{\sigma} \right| < 3 \tag{4}$$

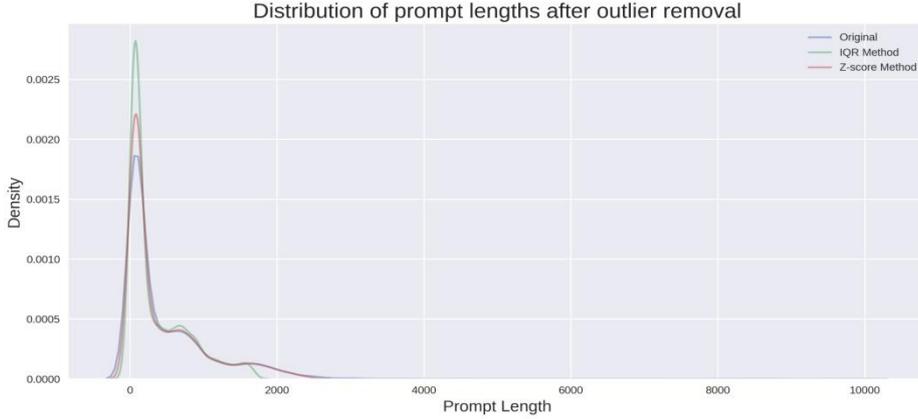

Fig.2. Distribution of prompt lengths show more prompts removed when removing outliers by IQR rather than z-score, which could be problematic for completeness of analysis.

The IQR and z-score methods vary in outlier sensitivity, particularly in long-tailed distributions, since z-score assumes normality, which this dataset is not. This normality assumption is likely to overly filter high-variance data [19], while the IQR method commonly suits skewed distributions [20]. As shown in the visualization that compares distribution by kernel density (Figure 3), z-score filtering was counterintuitively looser due to inflated $\sigma$ in the originally skewed distribution. IQR, being non-parametric, more aggressively excluded long-tail prompts. Since more variance was included in the z-score filtering, the longer tailed variant of the data made more sense to use in this analysis, to capture divergent characteristics of safety benchmarks [9], therefore the dataset after removing outliers by z-score was used in the analysis.

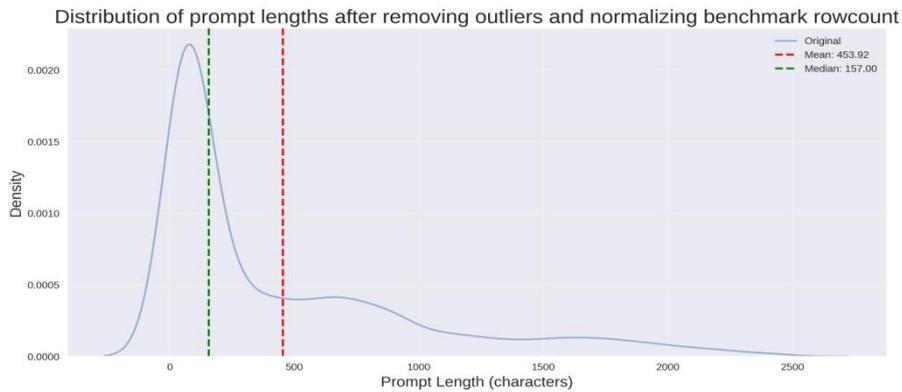

Fig.3. Distribution of prompt lengths showing a right-skewed density plot with a median of 157.0 characters (green dashed line) and mean of 453.92 characters (red dashed line) after outlier removal and benchmark normalization. The highest concentration lengths peak near 100 characters, with a long tail extending for a smaller number of prompts over 2500 characters.

## 3.2. Embedding Model Comparison

MiniLM and MPNet excel in different tasks, while similar in size, and were good candidates to use in this analysis for comparison [7] in a rigorous analysis. MiniLM provides fast, efficient embeddings with quality representation [21]. MPNet enhances contextual and sequential encoding, often surpassing BERT in NLP tasks [22]. While MiniLM emphasizes speed, MPNet enhances context modeling [7] and was used to determine the optimal silhouette score for differentiating embedding clusters.

## 3.3. Dimensionality Reduction Techniques

t-distributed Stochastic Neighbor Embedding (t-SNE) preserves local structure for fine-grained clustering but struggles with global relationships and is computationally greedy [23]. Uniform Manifold Approximation and Projection (UMAP) preserves both local and global structure while scaling efficiently, making it more suitable for larger datasets [24]. These methods complement each other; t-SNE is used for cluster separation, while UMAP handles larger data structures in AI safety semantic categorizations [25]. The selected embedding models normalize prompt value embeddings; however, UMAP can distort point distances [25] which is why these 2 methods are compared. The mathematical basis for normalization lies in ensuring consistent angle and gradient calculations throughout the pipeline [26]. While three-dimensional analysis is possible, we use two dimensions for optimal clustering capture without added noise [24], and we optimize by evaluating silhouette scores and computational efficiency for large-scale use.

## 3.4. Hyperparameter Selection within UMAP and T-SNE

Not all hyperparameters are essential [27], so we prioritized optimizing for maximum differences, enabling efficient replication by researchers. UMAP $n_{neighbors}$ balances local and global structures, while min dist maintains well-separated clusters [24], particularly with a min dist of 0.1 for tighter clusters. $n_{neighbors}$ was optimized due to the global and local structures that are evident in safety datasets [2], starting with a value of 15 and evaluating an increase to 30 [24]. For t-SNE, perplexity ensures robust neighborhood relationships and learning rate stabilizes embedding convergence [28]. These methods enhance cluster separation and embedding coherence, improving silhouette score differentiation. In optimizing t-SNE hyperparameters, we prioritized perplexity [28] while keeping the learning rate constant at 100, associating higher values with past optimization. Perplexity values from 5-50 are ideal, with a default recommendation of 30 [23], so we analyzed 30 and 50 using grid search, further optimizing for silhouette scores as well as computational efficiency.

## 3.5. Distance Metrics Compared

Distance metrics deterministically shape embedding similarity assessment and, consequently, cluster formation integrity. We compare Euclidean and Mahalanobis distances, each offering distinct semantic relationship perspectives in safety benchmark analysis. Euclidean distance, as calculated in Equation 5 below, works well in low-dimensional spaces with uniform variance but fails with high-dimensional correlated data, common in AI safety datasets [29].

$$d = \sqrt{(x_2 - x_1)^2 + (y_2 - y_1)^2} \qquad (5)$$

The Mahalanobis distance, as calculated in Equation 6, addresses these limitations by incorporating the embedding covariance matrix ($\Sigma$), establishing a statistical measure to account for dimensional correlations [30], which normalizes dimensional importance and accommodates anisotropic clusters [31]. This facilitates the inclusion of diverse semantic relationships, length variations, and potential semantic outliers, commonly seen in jailbreak attempts and adversarial prompting strategies [2].

$$d_M(x, y) = \sqrt{(x - y)^T \Sigma^{-1} (x - y)} \qquad (6)$$

Since the later distance metric considers covariance, where x and y are vectors, it has been proven effective in anisotropic clusters [30], while also relevant for longer prompts and outliers. Since these metrics complement each other, they were used as variables for comparison in the output [31].

## 3.6. Computational Efficiency Considerations

Computational efficiency measured via processing time [26] is a secondary metric that quantifies real-world scalability across the variants of this analysis. Our analysis was performed on a hosted Nvidia Tesla T4 GPU with 2,560 CUDA cores and 16 GB of GDDR6 memory, based on the Turing architecture, and the same hardware was used for all configurations. This efficiency analysis complements silhouette scores to identify configurations that balance semantic accuracy and computational feasibility [24], crucial for timely identification of self-harm patterns in evolving safety benchmarks, potentially accelerating intervention development for vulnerable populations.

## 3.7. Optimal Clustering Configurations Identified

Bootstrapping confidence intervals estimate uncertainty in silhouette scores [33], allowing comparison of multiple model configurations and highlighting the need to distinguish models with similar quality metrics by computational efficiency [26].

Dimensionality reduction optimizations for each variant is shown by the silhouette score strength $S_{sil}$ from [35], calculated in Equation 7 below, where $a(i)$ is the mean intra-cluster distance, while $b(i)$ is mean nearest-cluster distance for the point $i$.

$$S_{sil}(i) = \frac{b(i) - a(i)}{max(a(i), b(i))} \qquad (7)$$

As shown in Figure 4, MiniLM with Euclidean distance and UMAP reduction achieves one of the highest silhouette scores of $0.470 \pm 0.024$. Computationally, this configuration had the most efficient processing time (69.8s) among configurations showing statistically similar scores, making it the optimal clustering configuration [23]. This configuration's superiority aligns with earlier findings [7] that smaller, efficient transformer models like MiniLM can produce high-quality clustering embeddings despite reduced dimensionality, while also supporting research [24] indicating UMAP preserves both local and global structures better than t-SNE.

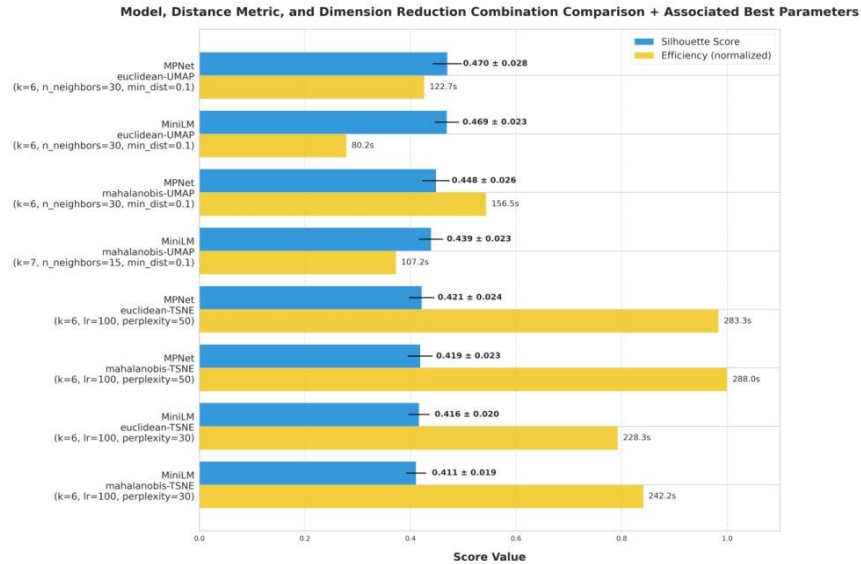

Fig.4. Comparative analysis of model configurations showing silhouette scores and computational efficiency in the form of processing time across different embedding models (MPNet, MiniLM), distance metrics (Euclidean, Mahalanobis), and dimension reduction techniques (UMAP, t-SNE).

## 3.8. Refinement for Optimum Number of Clusters

A higher sensitivity threshold refines the optimum cluster count by identifying groups that, while less distinct in the embedding space, represent unique semantic concepts [15]. Previous studies show that higher sensitivity thresholds reveal more significant structural changes in highdimensional data [34], so we analyzed this sensitivity using silhouette scores and confirmed it with the elbow method [36].

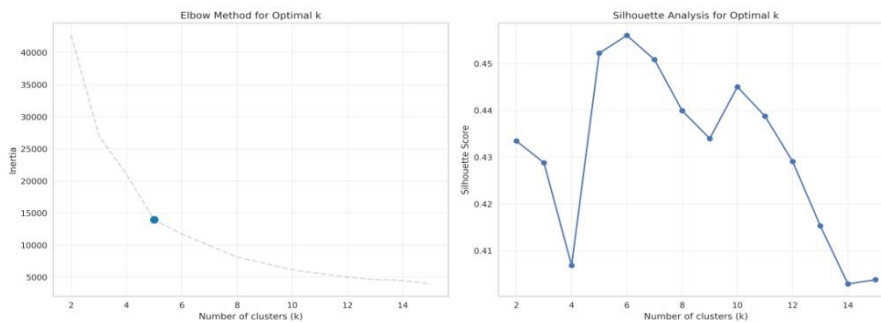

Fig.5. Cluster optimization analysis using elbow and silhouette methods. The elbow plot (left) indicates an optimal k=5 clusters, while the silhouette score analysis (right) shows peak performance at k=6 with a score of 0.455, which are close enough to validate each other.

While silhouette scores for each k provide insight into intra-cluster cohesion for individual data points, the elbow method focuses on the overall within-cluster sum of squares, calculated [36] in Equation 8 below, where $W_k$ is the sum of squares within clusters.

$$k_{opt} = \arg\min_k \{||\frac{\Delta W_k}{\Delta k} - \mu_{\Delta W}||\}$$

(8)

The convergence of both elbow and silhouette methods near k=6 (Figure 5) confirms that the semantic space of safety benchmarks naturally organizes into six distinct harm categories, supporting earlier findings [2] that optimal harm taxonomies typically consolidate into 5-7 primary categories despite more granular subcategorizations. We chose 6 since we expected a higher number of clusters from the taxonomy illustrated most recently in the Aegis 2.0 dataset [9], and best practice is to decide between possible clusters that best fit a use case [39].

## 4. CENTROID LABELING AND INFERENCE

We extracted four prompt values at each centroid edge based on past research [37] and created a prompt template using GPT-4 from OpenAI with the Aegis 2.0 taxonomy as labels [9]. We instructed the model to label anything outside the prompt values as 'Other' under the Aegis 2.0 taxonomy, adhering to best practices [37], but the model did not use this 'Other' definition. We ran this inference call multiple times to ensure consistent results, serving as bootstrapping for the labels found by model inference [33]. The five categories identified as most differentiated by the inference model were 'Hate/Identity Hate', 'Suicide and Self Harm', 'Guns/Illegal Weapons', 'PII/Privacy', and 'Criminal Planning'.

## 5. CLUSTERING RESULTS BY BENCHMARK

The UMAP dimensionality reduction visualization (Figure 6) revealed natural clustering of safety prompts into six distinct harm categories, with spatial organization that suggests underlying semantic relationships. The clear cluster separation (silhouette score 0.472, corroborated by an additional run) suggests strong thematic boundaries, while overlaps imply shared harm conceptualization among benchmark creators. This visualization emphasizes specific harm types, like GretelAI's focus on privacy and WildGuardMix's on self-harm scenarios. The spatial proximity between related concepts (e.g., Criminal Planning adjacent to Controlled Substances) further validates the embedding space's capture of semantic relationships. This clustering pattern provides empirical evidence for both the distinctness of harm categories and the varying coverage priorities across benchmark datasets, informing future benchmark development needs [2].

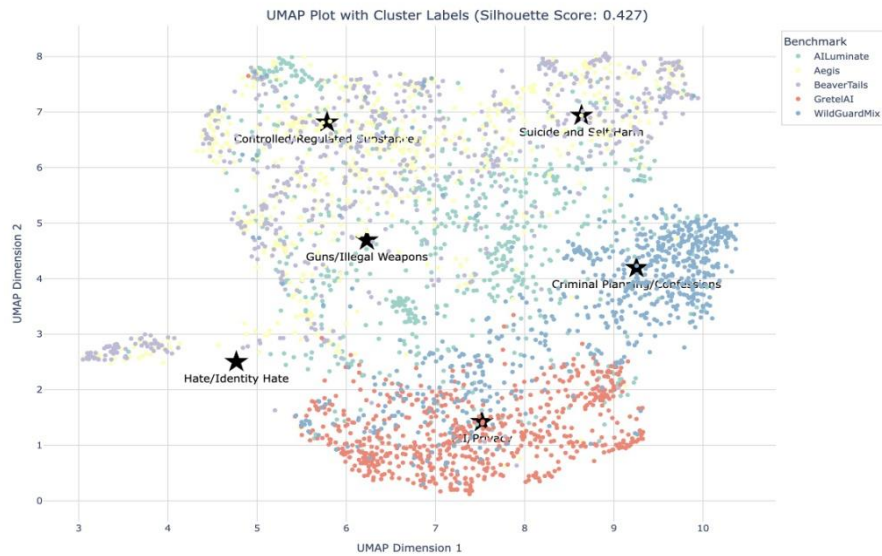

Fig.6. UMAP visualization of safety benchmark prompts colored by dataset source, with labeled clusters revealing distinct semantic groupings around key harm categories. The plot demonstrates clear separation between harm types while showing dataset-specific concentrations within certain domains.

## 6. CONTEXTUAL CONFOUNDS AND BIASES

The differences in density of prompt lengths from each dataset in semantic coverage around each cluster can be explained partially through biases in data collection. Prompt length distribution of the entire dataset used in analysis, for example (Figure 7), highlights potential differences in data collection strategies between benchmarks. In contrast to each other, GretelAI and WildGuardMix utilize substantially longer prompts (medians with over 700 characters), indicating more complex, context-rich scenarios for safety evaluation. This divergence indicates differing safety assessment approaches: shorter prompts target specific vulnerabilities, while longer prompts evaluate model behavior in nuanced real-world scenarios [38]. This divergence in prompt design likely reflects differing safety assessment approaches: shorter prompts target specific vulnerabilities, while longer ones evaluate model behavior in nuanced, real-world scenarios [38].

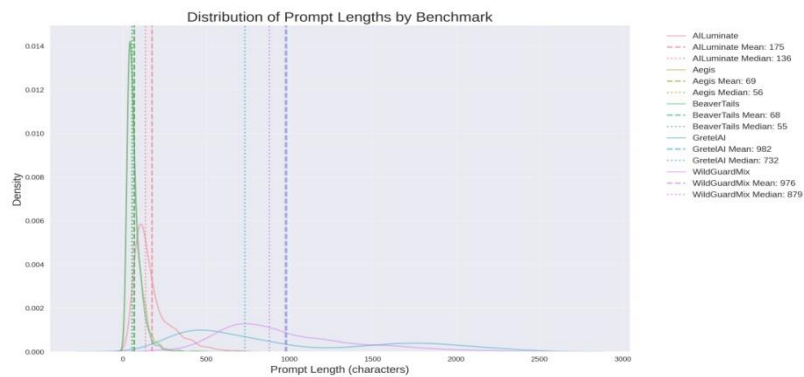

Fig.7. Kernel density distribution of prompt lengths across five AI safety benchmarks, revealing a bimodal pattern: Aegis, BeaverTails, and AILuminate favor concise prompts (medians 55-136 characters), while GretelAI and WildGuardMix employ substantially longer prompts (medians with over 700 characters).

# 7. LIMITATIONS

Our analysis has several methodological constraints. Using 85% confidence intervals ($\alpha$=0.15) rather than the conventional 95% [16] represents a deliberate trade-off between statistical rigor and practical detection sensitivity [39], introducing marginally higher Type I error potential while improving cluster detection. The clustering reveals patterns specific to these five benchmarks that may not generalize across broader AI safety contexts [1]. Our methodology exhibits potential biases, as dimensionality reduction loses information [25]; embedding model selection imposes representational constraints [7]; and implicit Western views of harm in the benchmarks [40] may skew category definitions. The ecological fallacy of generalizing cluster-level findings to individual prompts and the base rate fallacy of overemphasizing harm categories with lower prevalence further limit interpretation. Separate domain-specific safety datasets likely exhibit different semantic structures [2], and our equal benchmark weighting failed to reflect variations in deployment likelihood or importance. Future work should expand this framework to include benchmarks from more diverse cultural contexts [4], evaluate embedding bias propagation [15], and explore prompt-response relationships in adversarial settings [41]. Furthermore, we acknowledge potential disciplinary bias, as the authors' backgrounds predominantly in technology may limit our perspective on interdisciplinary trends and applications outside the technical domain, reflecting a homogeneity in expertise.

# 8. CONCLUSIONS

Our multidimensional clustering analysis identifies critical semantic gaps across AI safety benchmarks, revealing asymmetric coverage where WildGuardMix excels in self-harm detection while GretelAI prioritizes privacy over suicide content. These imbalances create vulnerabilities akin to social media, where content moderation flaws have led to self-harm proliferation. AI teams should conduct cross-benchmark evaluations using this dimensional reduction approach, focusing on underrepresented harm categories to proactively identify and mitigate emerging risks before deployment.

## ACKNOWLEDGMENTS

We thank Sari Andoni at Trase for guidance and MLCommons colleagues for their vital expertise in industry priorities, along with the reviewers of the International Conference on Advanced Natural Language Processing.